\documentclass{article}

\usepackage[preprint]{neurips_2025}

\usepackage{amsmath,amssymb,amsfonts,amsthm,mathtools,bm}
\usepackage{booktabs,microtype,xcolor,colortbl}
\usepackage{tikz}
\usetikzlibrary{arrows.meta,positioning}
\usepackage{pgfplots}
\pgfplotsset{compat=1.18}
\usepackage{mdframed,thmtools}
\definecolor{ink}{HTML}{1C2833}
\definecolor{accent}{HTML}{1A5276}
\definecolor{lane}{HTML}{EAF2F8}
\definecolor{lane2}{HTML}{FDEDEC}
\definecolor{good}{HTML}{196F3D}
\definecolor{warn}{HTML}{7B241C}
\definecolor{scholarblue}{rgb}{0.21,0.49,0.74}
\definecolor{rowhead}{HTML}{EAF2F8}
\usepackage{hyperref}
\hypersetup{colorlinks=true,linkcolor=scholarblue,citecolor=scholarblue,urlcolor=scholarblue}

\declaretheoremstyle[
  headfont=\sffamily\bfseries\color{accent},
  notefont=\sffamily\bfseries\color{accent},
  bodyfont=\normalfont\color{ink},
  spaceabove=2pt, spacebelow=2pt,
  mdframed={
    linewidth=0.45pt, linecolor=accent!40, backgroundcolor=accent!5,
    roundcorner=1.5pt, innertopmargin=2.5pt, innerbottommargin=2.5pt,
    innerleftmargin=6pt, innerrightmargin=6pt, skipabove=7pt, skipbelow=7pt
  }
]{soft}
\declaretheoremstyle[
  headfont=\sffamily\bfseries\color{accent},
  notefont=\sffamily\bfseries\color{accent},
  bodyfont=\normalfont,
  spaceabove=4pt, spacebelow=4pt
]{plainhead}
\declaretheoremstyle[
  headfont=\sffamily\bfseries\color{warn},
  notefont=\sffamily\bfseries\color{warn},
  bodyfont=\normalfont,
  spaceabove=4pt, spacebelow=4pt
]{warnhead}
\declaretheorem[style=soft,name=Theorem]{theorem}
\declaretheorem[style=soft,name=Proposition,sibling=theorem]{proposition}
\declaretheorem[style=soft,name=Corollary,sibling=theorem]{corollary}

\declaretheorem[style=plainhead,name=Definition,sibling=theorem]{definition}
\declaretheorem[style=plainhead,name=Assumption,sibling=theorem]{assumption}
\declaretheorem[style=warnhead,name=Remark,sibling=theorem]{remark}
\declaretheorem[style=plainhead,name=Example,sibling=theorem]{example}

\newcommand{\R}{\mathbb{R}}
\newcommand{\E}{\mathbb{E}}
\newcommand{\Prob}{\mathbb{P}}
\newcommand{\F}{\mathcal{F}}
\newcommand{\X}{\mathcal{X}}
\newcommand{\sg}{\operatorname{sg}}
\newcommand{\Cov}{\operatorname{Cov}}
\newcommand{\softmax}{\operatorname{softmax}}
\newcommand{\tr}{\operatorname{tr}}
\newcommand{\TV}{\operatorname{TV}}
\newcommand{\KL}{\operatorname{KL}}
\newcommand{\norm}[1]{\left\|#1\right\|}
\newcommand{\inner}[2]{\left\langle #1,#2\right\rangle}

\title{Predictive Latents for Autoregressive Language Models:\\
Possibility, Gauge, and Obstruction}

\author{%
  Yuzhou Hong \\
  Zhejiang Sci-Tech University \\
  \texttt{hongyuzhou@zstu.edu.cn}%
}

\begin{document}
\maketitle

\begin{abstract}
Autoregressive language models are trained by next-token cross-entropy, yet several recent systems also iterate a hidden state without emitting a token: continuous thoughts, pause steps, and recurrent depth.
This note separates three risks on one latent process.
The token risk $R_{\mathrm{tok}}$ scores a readout.
The teacher-forced latent risk $R_1$ scores a one-step map that sees the true hidden state.
The open-loop risk $R_H$ scores the same map after it has been iterated on its own outputs.
If the Bayes next-token posterior is a function of a $k$-dimensional statistic with full-rank covariance, a width-$k$ latent, a linear least-squares transition, and a refit unembedding match the Bayes token predictor.
The statistic is identified only up to an invertible gauge: whitening the residual stream does not change the token law once the unembedding is refit.
The matching obstructions are sharp.
A constant latent drives $R_1$ to zero while token cross-entropy falls back to the marginal entropy.
A perfect token predictor can still hide a null-space walk whose open-loop error grows linearly in the horizon.
A small next-latent loss does not certificate next-token prediction, and a small next-token loss does not certificate free latent rollout, unless a rank or injectivity condition is stated with the number.
\end{abstract}

\section{Introduction}
\label{sec:intro}

Next-token prediction is the training loss of modern language models~\citep{bengio2003nlm,vaswani2017attention,radford2019gpt2,brown2020gpt3,hoffmann2022chinchilla,touvron2023llama,grattafiori2024llama3}.
Read information-theoretically, that loss is compression of the token process~\citep{shannon1948theory,cover2006elements,deletang2024compression}.
The residual stream is where the computation is written~\citep{elhage2021framework,geva2022promotions,belrose2023tunedlens}.
Chain-of-thought spells intermediate steps as tokens~\citep{wei2022cot,kojima2022zeroshotcot}.
A newer family keeps the step inside the vector: continuous thoughts feed the hidden state back as the next embedding~\citep{hao2024coconut}, pause tokens and quiet rationales buy extra computation beside the answer~\citep{goyal2024pause,zelikman2024quietstar}, and recurrent-depth models unroll a latent block at test time~\citep{geiping2025recurrent,dehghani2019universal}.

\paragraph{Question.}
When is a next-latent map on an LLM hidden state a sufficient object for next-token prediction, and when does a small latent loss fail to control either the token distribution or a free rollout of the latent?

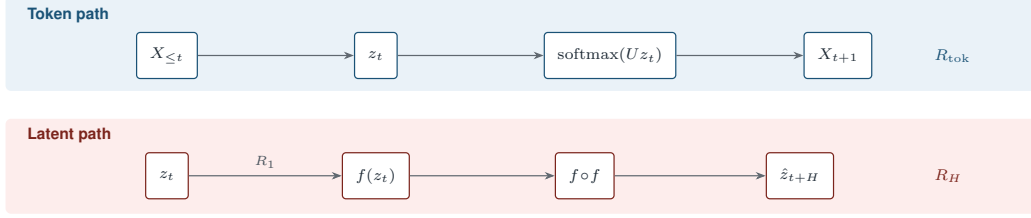
\begin{figure*}[t]
\centering
\resizebox{0.98\textwidth}{!}{%
\begin{tikzpicture}[
  font=\scriptsize,
  box/.style={draw=accent, rounded corners=1.6pt, fill=white, line width=0.55pt,
              minimum height=4.6ex, inner xsep=6pt, inner ysep=2pt, align=center, text=ink},
  arr/.style={-{Stealth[length=1.6mm]}, line width=0.55pt, draw=ink!70}
]
  \fill[lane, rounded corners=2.5pt] (0,0.22) rectangle (16.35,1.72);
  \fill[lane2, rounded corners=2.5pt] (0,-1.72) rectangle (16.35,-0.22);
  \node[anchor=west, font=\scriptsize\sffamily\bfseries, text=accent] at (0.22,1.42) {Token path};
  \node[box] (x) at (2.55,0.78) {$X_{\le t}$};
  \node[box] (z) at (5.85,0.78) {$z_t$};
  \node[box] (u) at (9.55,0.78) {$\softmax(Uz_t)$};
  \node[box] (tok) at (13.15,0.78) {$X_{t+1}$};
  \node[anchor=west, font=\scriptsize, text=accent] at (14.55,0.78) {$R_{\mathrm{tok}}$};
  \draw[arr] (x)--(z); \draw[arr] (z)--(u); \draw[arr] (u)--(tok);

  \node[anchor=west, font=\scriptsize\sffamily\bfseries, text=warn] at (0.22,-0.48) {Latent path};
  \node[box, draw=warn] (z0) at (2.55,-1.12) {$z_t$};
  \node[box, draw=warn] (f1) at (5.85,-1.12) {$f(z_t)$};
  \node[box, draw=warn] (f2) at (9.15,-1.12) {$f\!\circ\!f$};
  \node[box, draw=warn] (fH) at (12.55,-1.12) {$\hat z_{t+H}$};
  \node[anchor=west, font=\scriptsize, text=warn] at (14.55,-1.12) {$R_H$};
  \draw[arr] (z0)-- node[above, font=\tiny, text=ink!80] {$R_1$} (f1);
  \draw[arr] (f1)--(f2); \draw[arr] (f2)--(fH);
\end{tikzpicture}}
\caption{The two scores on one hidden state.
The top row reads $z_t$ out to the next token and never leaves the data prefix.
The bottom row iterates $f$ on its own outputs.
$R_1$ is only the first arrow of the bottom row.}
\label{fig:paths}
\end{figure*}

\paragraph{Contributions.}
(i) Token risk, teacher-forced latent risk, and open-loop latent risk are defined on one process (Section~\ref{sec:setup}).
(ii) A finite-dimensional predictive sufficient statistic is enough for Bayes token prediction, and linear least squares identifies it up to an invertible gauge (Section~\ref{sec:possible}).
(iii) Whitening the residual stream does not change the token law after the unembedding is refit (Section~\ref{sec:gauge}).
(iv) Two obstructions: latent collapse, and a null-space walk that the unembedding cannot see (Section~\ref{sec:obstruction}).
(v) A Lipschitz bound on free latent rollout, with a matching linear lower bound (Section~\ref{sec:horizon}).
Proofs are in the appendix.

\section{Related work}
\label{sec:related}

\paragraph{Autoregressive language models.}
Neural language models estimate a conditional distribution over the next token~\citep{bengio2003nlm,mikolov2010rnnlm}.
Transformers implement that conditional with self-attention~\citep{vaswani2017attention,dai2019transformerxl,su2024rope,dao2022flashattention}, and scale reliably with data and compute~\citep{kaplan2020scaling,hoffmann2022chinchilla,brown2020gpt3,touvron2023llama,touvron2023llama2,grattafiori2024llama3,jiang2023mistral,raffel2020t5}.
Bidirectional masked models are a different conditional~\citep{devlin2019bert}.
Instruction tuning and preference optimization change the target distribution, not the fact that the reported loss is a token loss~\citep{ouyang2022instructgpt,rafailov2023dpo}.
Analyses of the residual stream treat the hidden state as the place where the next-token computation is written~\citep{elhage2021framework,olsson2022induction,geva2021kv,geva2022promotions,belrose2023tunedlens,meng2022rome,todd2024functions}.
In-context learning has been explained as Bayesian inference and as implicit gradient descent on a latent task~\citep{xie2022bayesianicl,akyurek2023icl,vonoswald2023gd,saunshi2021mathematical}.
Those accounts already assume a latent that is sufficient for the next token.
This paper asks when a \emph{predictive} loss on that latent recovers the token conditional, and when it does not.

\paragraph{Predictive representations outside language.}
Joint-embedding architectures predict a future representation rather than pixels~\citep{lecun2022path,assran2023ijepa,bardes2024vjepa,assran2025vjepa2,balestriero2025lejepa,garrido2024learning}.
The same one-step idea appears for patch embeddings and for time series~\citep{xu2025nepa,chemeris2026lenepa}, and in masked, bootstrap, and contrastive training~\citep{he2022mae,baevski2022data2vec,grill2020byol,chen2020simclr,caron2021dino,oquab2024dinov2}.
Those objectives are context for the latent loss $R_1$.
The claims below are about the residual stream of a language model.

\paragraph{Latent computation inside language models.}
Chain-of-thought externalizes intermediate steps as tokens~\citep{wei2022cot,kojima2022zeroshotcot}.
Coconut keeps those steps in the residual stream and feeds the hidden state back as an embedding~\citep{hao2024coconut}.
Pause tokens and Quiet-STaR buy extra computation around the token channel~\citep{goyal2024pause,zelikman2024quietstar}.
Recurrent-depth models unroll a latent block for a variable number of steps at test time~\citep{geiping2025recurrent,dehghani2019universal}.
Each of these procedures is an open-loop or partially open-loop iteration on a vector.
The risk that training usually logs is still a token cross-entropy on the steps that do emit a symbol.
Section~\ref{sec:horizon} is the statement that those two numbers are different.

\paragraph{World models and imagined rollout.}
Latent imagination iterates a learned transition and scores a return, not a one-step residual~\citep{ha2018worldmodels,sutton1990dyna,hafner2020dreamer,hafner2021dreamerv2,hafner2023dreamerv3,schrittwieser2020muzero}.
The same gap appears when a language model iterates a hidden state.
The theorems below are the language-model form of that gap: no action channel is required for $R_1$ and $R_H$ to diverge.

\section{Setup}
\label{sec:setup}

Let $(X_t)_{t\in\mathbb{Z}}$ be a stationary process on a finite alphabet $\X$ with $|\X|=V<\infty$, and let $\F_t=\sigma(X_s:s\le t)$.
Write $\pi_t\in\Delta^{V-1}$ for the Bayes posterior,
\[
\pi_t(x)=\Prob(X_{t+1}=x\mid \F_t),
\]
and write $R_\star=\E[H(\pi_t)]=H(X_{t+1}\mid \F_t)$ for the Bayes token cross-entropy~\citep{shannon1948theory,cover2006elements}.

\begin{definition}[Latent and three risks]
\label{def:risks}
An encoder produces a latent $z_t\in\R^d$ that is $\F_t$-measurable.
A linear unembedding $U\in\R^{V\times d}$ induces
\[
R_{\mathrm{tok}}(U)=\E\bigl[-\log \softmax(Uz_t)_{X_{t+1}}\bigr].
\]
A measurable map $f:\R^d\to\R^d$ has teacher-forced latent risk
\[
R_1(f)=\E\norm{f(z_t)-z_{t+1}}^2
\]
and open-loop risk $R_H(f)=\E\norm{\hat z_{t+H}-z_{t+H}}^2$, where $\hat z_t=z_t$ and $\hat z_{t+k+1}=f(\hat z_{t+k})$.
\end{definition}

Teacher forcing feeds the encoded prefix.
Open-loop rollout feeds $f$ its own outputs.
Continuous thoughts and recurrent-depth steps are open-loop iterations of this kind on a subset of positions~\citep{hao2024coconut,geiping2025recurrent}.
Figure~\ref{fig:paths} draws the split.

\begin{assumption}[Moments]
\label{ass:moments}
$z_t$ is square-integrable, $R_1(f)<\infty$ for every $f$ under discussion, and $\Cov(z_t)$ exists.
\end{assumption}

\begin{remark}[What is not claimed]
\label{rem:scope}
We do not claim that $R_1$ should replace $R_{\mathrm{tok}}$ as a pretraining loss.
Next-token cross-entropy remains a consistent estimator of the conditional token law under standard conditions~\citep{bengio2003nlm,deletang2024compression}.
The claim is about certificates: which implications between $R_{\mathrm{tok}}$, $R_1$, and $R_H$ are valid.
\end{remark}

\section{Possibility}
\label{sec:possible}

A predictive latent is useful for an LLM when the next-token posterior is a function of a short vector, and when next-latent regression recovers that vector's transition.

\begin{definition}[Predictive sufficient statistic]
\label{def:pss}
A statistic $T_t\in\R^k$, measurable w.r.t.\ $\F_t$, is \emph{predictive sufficient} if there exists a measurable $h$ with $\pi_t=h(T_t)$ almost surely.
\end{definition}

This is sufficiency for the next token, not for the whole future $\sigma$-algebra.
It is the language-model form of a conditional predictive statistic: dimensions of the past that do not move $\pi_t$ are irrelevant to $R_{\mathrm{tok}}$~\citep{tishby2000ib,alemi2017vib,shwartz2017opening}.

\begin{theorem}[Width-$k$ possibility]
\label{thm:possible}
Suppose $T_t\in\R^k$ is predictive sufficient and $\Cov(T_t)$ is positive definite.
\begin{enumerate}
\item The encoder $z_t=T_t$ admits a (possibly nonlinear) readout with $R_{\mathrm{tok}}=R_\star$.
No encoder of any width can obtain a smaller token risk.
\item If $T_{t+1}=FT_t+\varepsilon_{t+1}$ with $\E[\varepsilon_{t+1}\mid\F_t]=0$ and $\E[\varepsilon_{t+1}\varepsilon_{t+1}^\top]=\Sigma$, the unique linear minimizer of $R_1$ is $f_\star(T)=FT$, and $R_1(f_\star)=\tr(\Sigma)$.
\item If $T'_t=GT_t$ for some $G\in\mathrm{GL}(k)$, then $T'$ is predictive sufficient, $\Cov(T')$ is positive definite, and the linear minimizer is $f'_\star(T')=GFG^{-1}T'$.
A readout of $T'$ that inverts $G$ and then applies $h$ attains $R_\star$.
\end{enumerate}
\end{theorem}

\paragraph{Sketch.}
Part (1) is the definition of sufficiency together with the fact that $\pi_t$ attains $R_\star$.
Part (2) is the normal equation for vector least squares; full rank of the covariance gives uniqueness.
Part (3) is the change-of-basis formula for that normal equation.
The appendix writes the algebra.

\begin{theorem}[Linear image of the posterior]
\label{thm:linearimage}
Suppose $z_t=A\pi_t$ and there exists $B$ such that $BA\pi=\pi$ for every $\pi$ in the simplex.
Then $z_t$ is predictive sufficient of dimension at most $d$, the readout $\pi_t=Bz_t$ attains $R_\star$, and any measurable readout of a non-injective image of $z_t$ cannot improve on $R_\star$.
\end{theorem}

Theorem~\ref{thm:linearimage} is the situation the tuned lens approximates empirically: an affine map from a hidden state to logits, refit per layer, recovers a token distribution that the residual stream already computes~\citep{belrose2023tunedlens,geva2022promotions,elhage2021framework}.
The theorem says when that recovery is exact.
It does not say that every layer of a trained transformer satisfies the hypothesis.

\begin{corollary}[Information lower bound]
\label{cor:ib}
If $I(z_t;X_{t+1})\le R_\star-\eta$ is impossible---more directly, if the Bayes risk of predicting $X_{t+1}$ from $z_t$ exceeds $R_\star$---then no readout of $z_t$ attains $R_\star$.
Equivalently, predictive sufficiency fails whenever $\pi_t$ is not $z_t$-measurable.
\end{corollary}

The corollary is the obstruction that an information bottleneck can impose on purpose~\citep{tishby2000ib,alemi2017vib}.
Compressing away coordinates that still move $\pi_t$ is not free.

\begin{example}[Bigram]
\label{ex:bigram}
Let $X_{t+1}$ be a function of $X_t$ only, on an alphabet of size $V$, with distinct rows of the transition matrix.
Then $T_t=X_t$ is predictive sufficient with $k\le V-1$ after a one-hot embedding, and $k=1$ is enough only if the rows lie on a line.
A next-latent loss on a $1$-dimensional projection is Bayes for the token problem if and only if that line condition holds.
\end{example}

\section{Gauge of the residual stream}
\label{sec:gauge}

A whitening, a rotation, or a stop-gradient on the target changes the coordinates of $z$ without changing the token conditional, once the unembedding is allowed to follow.

\begin{proposition}[Cosine is squared error on the sphere]
\label{prop:cosine}
If $\norm{u}=\norm{v}=1$, then $\tfrac12\norm{u-v}^2=1-\inner{u}{v}$.
On an encoder that already emits unit vectors, negative cosine similarity and $R_1$ have the same minimizers over $f$.
\end{proposition}

\begin{proposition}[Marginal penalties do not select $f$]
\label{prop:sg}
Fix the law of $(z_t,z_{t+1})$.
Over maps $f$, $\E\norm{f(z_t)-\sg(z_{t+1})}^2$ has the same minimizers as $R_1$.
The penalty $\norm{\Cov(z_t)-I}_F^2$ does not depend on $f$.
\end{proposition}

Stop-gradient matters for the encoder update, where it blocks a symmetric collapse~\citep{grill2020byol,chen2021simsiam}.
It is not an extra constraint on the transition once embeddings are treated as data.
A penalty on $\Cov(z_t)$ is a constraint on the marginal of the residual stream~\citep{balestriero2025lejepa}.

\begin{theorem}[Unembedding gauge]
\label{thm:gauge}
Let $G\in\mathrm{GL}(d)$, $z'_t=Gz_t$, and $U'=UG^{-1}$.
Then $\softmax(U'z'_t)=\softmax(Uz_t)$ almost surely, so $R_{\mathrm{tok}}(U')=R_{\mathrm{tok}}(U)$.
If $f$ minimizes $R_1$ for $z$ in a linear class, the conjugate $f'(z')=Gf(G^{-1}z')$ minimizes $R_1$ for $z'$.
\end{theorem}

Thus a whitening of the residual stream, or an isotropy penalty that is exactly realized by an invertible reparameterization, does not change the token predictor after the unembedding is refit.
Reporting a change in $R_1$ under a gauge change, without refitting $U$, does not measure a change in $R_{\mathrm{tok}}$.

\section{Obstructions}
\label{sec:obstruction}

\begin{theorem}[Collapse]
\label{thm:collapse}
Suppose $z_t=c$ almost surely for a constant $c\in\R^d$.
Then $f\equiv c$ attains $R_1(f)=0$ and $R_H(f)=0$.
Every readout of $z_t$ satisfies $R_{\mathrm{tok}}\ge H(X_{t+1})$, with equality at the marginal distribution of $X_{t+1}$.
The gap $H(X_{t+1})-R_\star$ equals $I(X_{t+1};\F_t)$ and can be arbitrarily close to $\log V$ while $R_1=0$.
\end{theorem}

Constant embeddings are minimizers of symmetric predictive losses unless an asymmetric constraint is imposed~\citep{grill2020byol,chen2021simsiam,balestriero2025lejepa}.
Theorem~\ref{thm:collapse} records the token-level cost of that minimizer: the latent certificate is perfect and the language model is an unconditional unigram.

\begin{theorem}[Non-measurable posterior]
\label{thm:nonmeas}
If $\pi_t$ is not almost surely a function of $z_t$, then every measurable readout of $z_t$ has token risk strictly above $R_\star$.
The excess risk is $\E\bigl[\KL(\pi_t\,\|\,\bar\pi(z_t))\bigr]$, where $\bar\pi(z)=\E[\pi_t\mid z_t=z]$.
\end{theorem}

Histories that the encoder ties together cannot be untied by a deeper unembedding.
This is the LLM form of an encoder that drops a control-relevant bit: here the ``control'' is the next-token posterior.
Probes that only read a collapsed coordinate inherit the excess KL~\citep{belrose2023tunedlens,saunshi2021mathematical}.

\begin{theorem}[Token Bayes with a bad rollout]
\label{thm:null}
Let $z_t=(T_t,s_t)$ with $T_t$ predictive sufficient, $s_{t+1}=s_t+\varepsilon_{t+1}$, $\E[\varepsilon_{t+1}\mid\F_t]=0$, $\E\norm{\varepsilon_{t+1}}^2=\rho$, and $\varepsilon_{t+1}$ independent of $T$.
Let $U$ ignore the $s$-block, and let $f(T,s)=(F_\star T,\, s)$ where $F_\star$ is the linear minimizer on $T$ from Theorem~\ref{thm:possible}.
Then a readout of $T_t$ attains $R_{\mathrm{tok}}=R_\star$, while the open-loop risk of this $f$ on the whole latent satisfies $R_H(f)=H\rho$.
\end{theorem}

A language model can be Bayes for the next token and still drift in a null direction of the unembedding when the hidden state is iterated.
Free latent generation---continuous thoughts, extra recurrent depth---sees that direction~\citep{hao2024coconut,geiping2025recurrent}.
Token cross-entropy on teacher-forced positions does not.

\section{Open-loop horizon}
\label{sec:horizon}

\begin{assumption}[Lipschitz transition]
\label{ass:lip}
$\norm{f(z)-f(z')}\le L\norm{z-z'}$ for all $z,z'$.
\end{assumption}

\begin{theorem}[Propagation]
\label{thm:lip}
Under Assumptions~\ref{ass:moments} and~\ref{ass:lip},
\begin{equation}
\label{eq:lip}
\sqrt{R_{k+1}(f)}
\le L\sqrt{R_k(f)}+\sqrt{\E\norm{f(z_{t+k})-z_{t+k+1}}^2}.
\end{equation}
If the one-step residual is at most $\rho$ at every time,
\[
\sqrt{R_H(f)}
\le
\begin{cases}
H\sqrt{\rho}, & L=1,\\
\sqrt{\rho}\,\dfrac{L^H-1}{L-1}, & L\neq 1.
\end{cases}
\]
If $L<1$, then $\sup_H R_H(f)\le \rho/(1-L)^2$.
\end{theorem}

\begin{theorem}[Linear lower bound]
\label{thm:lower}
There exist a latent process and a $1$-Lipschitz $f$ with $R_1(f)=\rho$ and $R_H(f)=H\rho$ for every $H\ge 1$.
The factor $H$ in Theorem~\ref{thm:lip} is therefore sharp at $L=1$.
\end{theorem}

\begin{proposition}[From latent error to token total variation]
\label{prop:tv}
Suppose $\softmax(Uz)$ is $\ell$-Lipschitz from $(\R^d,\norm{\cdot})$ into $(\Delta,\norm{\cdot}_1)$.
Then the expected total variation between the teacher-forced token law $\softmax(Uz_{t+H})$ and the open-loop law $\softmax(U\hat z_{t+H})$ is at most $(\ell/2)\sqrt{R_H(f)}$.
\end{proposition}

On any set where the logits $Uz$ stay bounded, softmax is Lipschitz, so the proposition applies to a residual stream that is not exploding~\citep{elhage2021framework,khalil2002nonlinear}.
It does not apply uniformly on all of $\R^d$: softmax saturates, and a large latent error can move probability by at most $1$.

\paragraph{Where language-model procedures sit.}
Teacher-forced next-token training scores $R_{\mathrm{tok}}$ on the true prefix~\citep{radford2019gpt2,brown2020gpt3}.
Continuous thoughts and recurrent depth iterate the latent and only sometimes decode~\citep{hao2024coconut,geiping2025recurrent}.
Theorem~\ref{thm:lip} says the decoded law after $H$ latent steps is controlled by $R_1$ only through the Lipschitz factor of $f$.
If that factor is near $1$, a horizon of the length used for continuous thoughts multiplies the one-step residual.
If the factor exceeds $1$, the bound is exponential, and Theorem~\ref{thm:lower}'s linear example shows that one cannot remove the horizon factor in general.

\begin{figure}[t]
\centering
\begin{tikzpicture}
\begin{axis}[
  width=\linewidth,
  height=0.72\linewidth,
  xlabel={horizon $H$},
  ylabel={open-loop risk $R_H$},
  xmin=1, xmax=16,
  ymin=0, ymax=16.4,
  xtick={1,4,8,12,16},
  tick label style={font=\scriptsize, color=ink},
  label style={font=\scriptsize, color=ink},
  legend style={font=\scriptsize, draw=accent!30, fill=white, at={(0.97,0.03)}, anchor=south east},
  axis line style={draw=ink!55},
  tick style={draw=ink!45},
  every axis plot/.append style={line width=1.05pt},
]
\addplot[accent, domain=1:16, samples=32] {(1-0.9^(2*x))/(1-0.81)};
\addplot[warn, domain=1:16, samples=2] {x};
\addplot[good, densely dotted, domain=1:16, samples=2] {1/0.19};
\legend{$a=0.9$ optimal, $a=1$ walk, ceiling}
\end{axis}
\end{tikzpicture}
\caption{Open-loop risk when the one-step residual is already $1$.
The optimal linear map ($a=0.9$) rises toward $5.26$.
The unit-Lipschitz walk ($a=1$) keeps growing.}
\label{fig:horizon}
\end{figure}
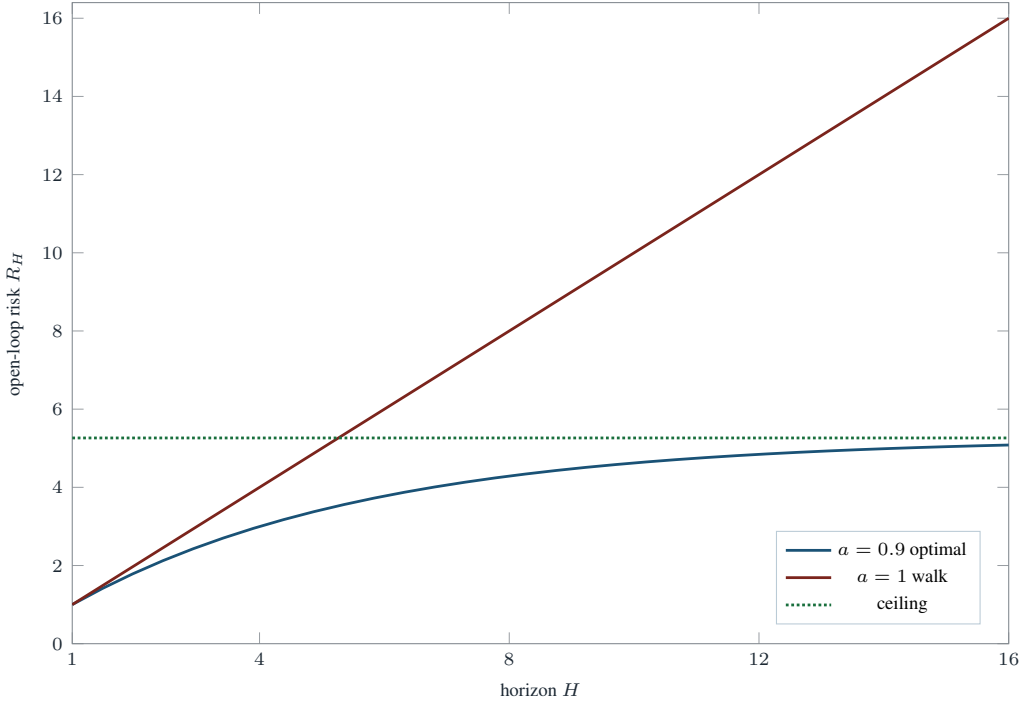

\begin{example}[Scalar autoregressive latent]
\label{ex:ar}
Take $z_{t+1}=a z_t+\varepsilon_{t+1}$ with $|a|<1$, $\E[\varepsilon_{t+1}\mid\F_t]=0$, and $\E[\varepsilon_{t+1}^2]=\sigma^2$.
The least-squares map is $f_\star(z)=az$, and $R_1(f_\star)=\sigma^2$.
The open-loop error is the accumulated innovation $\sum_{j=0}^{H-1}a^{j}\varepsilon_{t+H-j}$, so
\[
R_H(f_\star)=\sigma^2\frac{1-a^{2H}}{1-a^2}.
\]
For $a=0.9$ and $\sigma^2=1$, one has $R_1=1$, $R_8\approx 4.29$, and $R_{16}\approx 5.08$, while the stationary ceiling is $\sigma^2/(1-a^2)\approx 5.26$.
Figure~\ref{fig:horizon} plots that curve against the unit-Lipschitz walk.
The one-step residual is already optimal.
The growth is the pushed-forward noise, not a training failure.
If instead $f(z)=z$ (the unit-Lipschitz predictor of Theorem~\ref{thm:lower}) and $a=1$, then $R_H=H\sigma^2$, so eight latent steps cost eight times the one-step residual.
A continuous-thought block of that length~\citep{hao2024coconut} sits in this range: teacher-forced $R_1$ understates the open-loop displacement by the factor $(1-a^{2H})/(1-a^2)$.
\end{example}

\begin{table}[t]
\centering
\caption{Which small number implies which conclusion.
The side condition is required for the implication.}
\label{tab:cert}
\renewcommand{\arraystretch}{1.15}
\begin{tabular}{lll}
\toprule
\rowcolor{rowhead}
\textbf{Small number} & \textbf{Conclusion} & \textbf{Side condition} \\
\midrule
$R_1$ & $R_{\mathrm{tok}}$ near $R_\star$ & sufficiency, Thm.~\ref{thm:possible} \\
$R_1=0$ & token Bayes & fails under collapse, Thm.~\ref{thm:collapse} \\
\rowcolor{lane2}
$R_{\mathrm{tok}}=R_\star$ & small $R_H$ & fails in a null direction, Thm.~\ref{thm:null} \\
$R_1$ & small $R_H$ for all $H$ & $L<1$, Thm.~\ref{thm:lip} \\
\rowcolor{lane}
isotropy of $z$ & a new token law & none; it is a gauge, Thm.~\ref{thm:gauge} \\
\bottomrule
\end{tabular}
\end{table}

\section{What the theorems allow}
\label{sec:allow}

\paragraph{A latent can replace the token loss.}
If a practitioner exhibits a $k$-dimensional $T_t$ with $\pi_t=h(T_t)$ and a full-rank covariance, Theorem~\ref{thm:possible} allows them to train a linear next-latent map and a readout, and to claim Bayes token risk.
The integer $k$ is part of the claim.
Hidden-state probes that recover the next token are evidence toward the hypothesis, not a proof that it holds for the trained network~\citep{belrose2023tunedlens,geva2022promotions}.

\paragraph{A latent loss is a token loss only after a readout exists.}
On the sphere, Proposition~\ref{prop:cosine} identifies cosine similarity with $R_1$.
Without a map from the latent to a distribution over the vocabulary, $R_1$ is not a language-model loss.
Attaching a linear head and invoking Theorem~\ref{thm:linearimage} is the step that makes it one.
A covariance penalty survives that step only as a gauge (Theorem~\ref{thm:gauge}, Proposition~\ref{prop:sg}).

\paragraph{Free latent steps need an open-loop number.}
A continuous-thought block of length $H$~\citep{hao2024coconut} or a recurrent depth $H$~\citep{geiping2025recurrent} is an open-loop trajectory.
Table~\ref{tab:cert} says a teacher-forced token loss on the symbols before and after the block does not bound the latent drift inside the block.
The bound that does apply is Theorem~\ref{thm:lip}, and it needs $L$ or a measured $R_H$.

\paragraph{Compression is allowed only along the kernel of $\pi_t$.}
Coordinates of the past that do not move the next-token posterior may be dropped (Definition~\ref{def:pss}, Corollary~\ref{cor:ib}).
Coordinates that do move it may not (Theorem~\ref{thm:nonmeas}).
An information-bottleneck penalty is compatible with Bayes token prediction only when its bottleneck leaves a predictive sufficient statistic intact~\citep{tishby2000ib,alemi2017vib,shwartz2017opening}.

\section{Discussion}
\label{sec:disc}

\paragraph{Squared loss versus cross-entropy.}
$R_1$ is a squared residual because that is the population criterion whose minimizer is the conditional mean of $z_{t+1}$.
Token risk is a cross-entropy because the observation is discrete.
Proposition~\ref{prop:cosine} is the bridge between a cosine comparison and squared error: after normalization they agree.
Nothing here extends automatically to a KL between full conditionals over future strings.
Multi-token predictive sufficiency is stronger than one-step sufficiency and is not implied by Theorem~\ref{thm:possible}.

\paragraph{Nonlinear transitions.}
Part (2) of Theorem~\ref{thm:possible} uses a linear $f$.
For a nonlinear conditional mean, $R_1$ still selects $\E[z_{t+1}\mid z_t]$ in an unrestricted nonparametric class, but the open-loop iterate of that mean is not the multi-step conditional mean.
Theorem~\ref{thm:lip} is the replacement bound: it needs Lipschitz regularity and does not restore equality~\citep{khalil2002nonlinear}.

\paragraph{Context length.}
The encoder may read a window rather than the full past, as in transformer language models with a finite context~\citep{vaswani2017attention,dai2019transformerxl}.
Predictive sufficiency is then sufficiency relative to that window.
If the Bayes posterior depends on tokens outside the window, Theorem~\ref{thm:nonmeas} applies with $z_t$ equal to the window state.

\paragraph{Actions.}
We did not put an action channel in $f$.
Standard language-model pretraining has none~\citep{radford2019gpt2}.
Tool use adds one; the horizon statements extend verbatim to an action-conditioned Lipschitz map, with $R_1$ evaluated along the action sequence.
That extension is not needed for the obstructions in Section~\ref{sec:obstruction}.

\section{Conclusion}
\label{sec:conc}

Next-token language models and next-latent predictors share a prefix and a one-step arrow, and they score different objects.
A finite-dimensional predictive statistic is enough for both: linear least squares recovers its transition up to gauge, and a refit unembedding recovers the Bayes token law.
The same diagram contains the failures.
A collapsed latent makes $R_1$ vanish while token cross-entropy falls back to the marginal.
A Bayes token predictor can still wander in the kernel of the unembedding, so free latent rollout---continuous thought, recurrent depth---is not certified by teacher-forced perplexity~\citep{hao2024coconut,geiping2025recurrent}.
The certificate that matches the claim is the pair $(R_{\mathrm{tok}}, R_H)$, together with the rank of the latent that is alleged to be sufficient.

\bibliography{refs}

\appendix
\section{Proofs}
\label{app:proofs}

\subsection{Theorem~\ref{thm:possible}}
\label{app:possible}

\paragraph{Part (1).}
By Definition~\ref{def:pss}, $\pi_t=h(T_t)$.
The predictor that outputs $h(T_t)$ has conditional cross-entropy $H(\pi_t)$ at each time, hence risk $R_\star$.
For any encoder and any readout, the conditional law of $X_{t+1}$ given the latent is a mean of $\pi_t$ (or a garbling of it).
Cross-entropy is at least the entropy of that conditional law, so the risk is at least $R_\star$~\citep{cover2006elements}.

\paragraph{Part (2).}
Consider $f(T)=WT$ with $W\in\R^{k\times k}$.
Expand
\[
R_1(W)=\E\norm{WT_t-T_{t+1}}^2
=\tr\bigl(W C_0 W^\top - W C_1^\top - C_1 W^\top + C_+\bigr),
\]
where $C_0=\E[T_t T_t^\top]$, $C_1=\E[T_{t+1}T_t^\top]$, and $C_+=\E[T_{t+1}T_{t+1}^\top]$.
The gradient in $W$ vanishes at $W C_0=C_1$.
Positive-definiteness of $C_0=\Cov(T_t)+\E[T_t]\E[T_t]^\top$ follows from positive-definiteness of the covariance after centering; if the mean is nonzero, $C_0$ remains positive definite whenever $\Cov(T_t)$ is.
Thus $W_\star=C_1 C_0^{-1}$ is the unique critical point and the unique minimizer.
Under $T_{t+1}=FT_t+\varepsilon_{t+1}$ with $\E[\varepsilon_{t+1}\mid\F_t]=0$, one has $C_1=F C_0$, so $W_\star=F$.
The residual is $\varepsilon_{t+1}$, and $R_1(f_\star)=\E\norm{\varepsilon_{t+1}}^2=\tr(\Sigma)$.

\paragraph{Part (3).}
$T'_t=GT_t$ with $G$ invertible.
Then $\pi_t=h(G^{-1}T'_t)$, so $T'$ is predictive sufficient.
$\Cov(T'_t)=G\Cov(T_t)G^\top$ is positive definite.
The population minimizer transforms as
\[
W'_\star
=\E[T'_{t+1}{T'_t}^\top]\E[T'_t{T'_t}^\top]^{-1}
= (G F C_0 G^\top)(G C_0 G^\top)^{-1}
= G F G^{-1},
\]
which is the claim.
The readout $h(G^{-1}T'_t)$ attains $R_\star$ by Part (1).

\subsection{Theorem~\ref{thm:linearimage}}
\label{app:linearimage}

$BA\pi_t=\pi_t$ shows that $\pi_t$ is a function of $z_t=A\pi_t$, so $z_t$ is predictive sufficient and the linear readout $Bz_t$ attains $R_\star$.
No readout attains a smaller risk, by the same argument as Theorem~\ref{thm:possible}(1).

\subsection{Corollary~\ref{cor:ib}}
\label{app:ib}

The Bayes risk of predicting $X_{t+1}$ from $z_t$ equals $\E[H(\bar\pi(z_t))]$, where $\bar\pi(z)=\E[\pi_t\mid z_t=z]$.
This quantity equals $R_\star$ if and only if $\pi_t=\bar\pi(z_t)$ almost surely, i.e., $\pi_t$ is $z_t$-measurable.
Otherwise the risk of every readout is at least $\E[H(\bar\pi(z_t))]>R_\star$.

\subsection{Proposition~\ref{prop:cosine}}
\label{app:cosine}

For unit vectors,
\[
\norm{u-v}^2=\norm{u}^2+\norm{v}^2-2\inner{u}{v}=2-2\inner{u}{v}.
\]
Dividing by $2$ gives the identity.
The two losses differ by the additive constant $1$ on the unit sphere, so they have the same argmin over $f$.

\subsection{Proposition~\ref{prop:sg}}
\label{app:sg}

Stop-gradient does not change the value of the random variable $z_{t+1}$ inside the expectation when the law of $(z_t,z_{t+1})$ is held fixed.
The two criteria are therefore the same function of $f$.
The Frobenius penalty depends on the law of $z_t$ only.

\subsection{Theorem~\ref{thm:gauge}}
\label{app:gauge}

$U'z'_t=UG^{-1}Gz_t=Uz_t$, so the logits agree and so do the softmax distributions and the token risks.
For the transition, substitute $z=G^{-1}z'$ into a linear minimizer $f(z)=Wz$ of the original problem.
The conjugate $f'(z')=G W G^{-1} z'$ is exactly the minimizer produced by the change-of-basis computation in Theorem~\ref{thm:possible}(3).

\subsection{Theorem~\ref{thm:collapse}}
\label{app:collapse}

If $z_t=z_{t+1}=c$, then $f\equiv c$ gives residual $0$, and the open-loop trajectory stays at $c$, so $R_H=0$.
A readout of a constant latent is a single distribution $p\in\Delta^{V-1}$.
Its risk is $\E[-\log p(X_{t+1})]=H(X_{t+1})+\KL(\mathrm{Law}(X_{t+1})\,\|\,p)$, which is minimized at the marginal and then equals $H(X_{t+1})$.
The identity $H(X_{t+1})-H(X_{t+1}\mid\F_t)=I(X_{t+1};\F_t)$ is the chain rule~\citep{cover2006elements}.
For a deterministic bigram on $V$ symbols with uniform stationary distribution, $R_\star=0$ and $H(X_{t+1})=\log V$.

\subsection{Theorem~\ref{thm:nonmeas}}
\label{app:nonmeas}

Any readout induces a conditional distribution $q(\cdot\mid z_t)$.
Taking the expectation of the cross-entropy and using the conditional Jensen inequality for the convex map $p\mapsto -\log q$ gives
\[
R_{\mathrm{tok}}(q)
\ge \E\bigl[-\textstyle\sum_x \pi_t(x)\log q(x\mid z_t)\bigr]
\ge \E\bigl[-\textstyle\sum_x \bar\pi(z_t)(x)\log \bar\pi(z_t)(x)\bigr],
\]
and the gap from $\E[H(\pi_t)]$ to the middle expression, when $q=\bar\pi$, equals $\E[\KL(\pi_t\,\|\,\bar\pi(z_t))]$.
The KL is strictly positive on a set of positive probability precisely when $\pi_t$ is not $\bar\pi(z_t)$ almost surely, because $\KL(p\|q)=0$ if and only if $p=q$.

\subsection{Theorem~\ref{thm:null}}
\label{app:null}

The $T$-block is predictive sufficient, so a readout of $T_t$ attains $R_\star$ by Theorem~\ref{thm:possible}(1), and $U$ that ignores $s$ can realize that readout on the first block.
The open-loop map leaves $s$ unchanged: $\hat s_{t+H}=s_t$, while
\[
s_{t+H}=s_t+\sum_{j=1}^{H}\varepsilon_{t+j}.
\]
Independence of the increments and $\E\norm{\varepsilon}^2=\rho$ give
\[
\E\norm{\hat z_{t+H}-z_{t+H}}^2
=\E\Bigl\|\sum_{j=1}^{H}\varepsilon_{t+j}\Bigr\|^2
=H\rho,
\]
where the $T$-block contributes $0$ because $f$ uses the true linear minimizer and, in the construction, we evaluate the $s$-component only.
If the $T$-block is itself noisy, its contribution is nonnegative and the lower bound $H\rho$ still holds for the $s$-component alone.
The statement in the main text uses a noise-free $T$ transition so that the whole-vector risk equals $H\rho$.

\subsection{Theorem~\ref{thm:lip}}
\label{app:lip}

Write the error $e_k=\hat z_{t+k}-z_{t+k}$.
Then
\begin{align*}
e_{k+1}
&=f(\hat z_{t+k})-z_{t+k+1}\\
&=f(\hat z_{t+k})-f(z_{t+k})+f(z_{t+k})-z_{t+k+1}.
\end{align*}
Lipschitz continuity yields $\norm{f(\hat z_{t+k})-f(z_{t+k})}\le L\norm{e_k}$.
Minkowski's inequality in $L^2$ gives
\[
\sqrt{\E\norm{e_{k+1}}^2}
\le L\sqrt{\E\norm{e_k}^2}+\sqrt{\E\norm{f(z_{t+k})-z_{t+k+1}}^2},
\]
which is~\eqref{eq:lip}.
Let $\rho_k=\E\norm{f(z_{t+k})-z_{t+k+1}}^2\le\rho$ and $a_k=\sqrt{R_k(f)}$, with $a_0=0$ because $\hat z_t=z_t$.
Then $a_{k+1}\le L a_k+\sqrt{\rho}$.
Unrolling,
\[
a_H\le \sqrt{\rho}\sum_{j=0}^{H-1} L^{j},
\]
which is the closed form in the statement.
If $L<1$, the infinite sum is $1/(1-L)$, so $R_H\le \rho/(1-L)^2$.

\subsection{Theorem~\ref{thm:lower}}
\label{app:lower}

Take $d=1$, $z_{t+1}=z_t+\varepsilon_{t+1}$ with $\E[\varepsilon_{t+1}\mid\F_t]=0$, $\E[\varepsilon_{t+1}^2]=\rho$, and independent increments.
The map $f(z)=z$ is $1$-Lipschitz and has $R_1(f)=\rho$.
The same calculation as in Appendix~\ref{app:null} gives $R_H(f)=H\rho$.

\subsection{Proposition~\ref{prop:tv}}
\label{app:tv}

Let $p=\softmax(Uz_{t+H})$ and $\hat p=\softmax(U\hat z_{t+H})$.
Lipschitz continuity gives $\norm{p-\hat p}_1\le \ell\norm{z_{t+H}-\hat z_{t+H}}$.
Total variation is half the $\ell_1$ distance, so
\[
\E[\TV(p,\hat p)]
\le \frac{\ell}{2}\E\norm{z_{t+H}-\hat z_{t+H}}
\le \frac{\ell}{2}\sqrt{\E\norm{z_{t+H}-\hat z_{t+H}}^2}
=\frac{\ell}{2}\sqrt{R_H(f)},
\]
where the second step is Jensen's inequality for the square root, or equivalently Cauchy--Schwarz on the expectation of the norm.

\subsection{A note on the sphere constraint inside $R_1$}
\label{app:sphere}

If the encoder is constrained to the unit sphere, the conditional least-squares problem of Theorem~\ref{thm:possible}(2) is replaced by a constrained problem.
Proposition~\ref{prop:cosine} says the cosine objective equals that constrained squared loss.
The gauge group shrinks from $\mathrm{GL}(d)$ to the orthogonal group, because only orthogonal maps preserve the sphere and the Euclidean residual simultaneously.
Theorem~\ref{thm:gauge} still applies to every $G$ in that subgroup: the unembedding absorbs $G^{-1}$.
Isotropy penalties whose minimizers are orthogonal whitenings sit inside this subgroup when the latent is already full rank~\citep{balestriero2025lejepa,chemeris2026lenepa}.

\end{document}